\documentclass{article}

\usepackage{arxiv}

\usepackage[utf8]{inputenc} 
\usepackage[T1]{fontenc}    
\usepackage{hyperref}       
\usepackage{url}            
\usepackage{booktabs}       
\usepackage{amsfonts}       
\usepackage{nicefrac}       
\usepackage{microtype}      
\usepackage{lipsum}
\usepackage{graphicx}
\graphicspath{ {./images/} }

\usepackage{epsfig} 
\usepackage{mathptmx} 
\usepackage{times} 
\usepackage{amsmath} 
\usepackage{amssymb}  
\usepackage{cite}
 \usepackage[table,xcdraw]{xcolor}

\title{3D Roadway Scene Object Detection with LiDARs in Snowfall Conditions}

\author{
  Ghazal Farhani\thanks{This is an extended version of a paper published in 
  \textit{2024 IEEE 27th International Conference on Intelligent 
  Transportation Systems (ITSC)}, pp. 1441--1448, Sept. 2024. 
  \copyright~2024 IEEE. Personal use permitted.} \\
  National Research Council Canada, London, Ontario, Canada\\
  \texttt{ghazal.farhani@nrc-cnrc.gc.ca}
  \And
  Taufiq Rahman \\
  National Research Council Canada, London, Ontario, Canada\\
  \texttt{Taufiq.Rahman@nrc-cnrc.gc.ca}
  \And
  Syed Mostaquim Ali \\
  Western University and National Research Council Canada, London, Ontario, Canada\\
  \texttt{syedmostaquim.ali@nrc-cnrc.gc.ca}
  \And
  Andrew Liu\thanks{Work performed while at National Research Council Canada.} \\
  \texttt{andrewmh.liu@gmail.com}
  \And
  Mohamed Zaki \\
  Western University, London, Ontario, Canada\\
  \texttt{mzaki9@uwo.ca}
  \And
  Dominique Charlebois \\
  Transport Canada, Ottawa, Ontario, Canada\\
  \texttt{dominique.charlebois@tc.gc.ca}
  \And
  Benoit Anctil \\
  Transport Canada, Ottawa, Ontario, Canada\\
  \texttt{benoit.anctil@tc.gc.ca}
}

\begin{document}
\maketitle
\begin{abstract}
Because 3D structure of a roadway environment can be characterized directly by a Light Detection and Ranging (LiDAR) sensors, they can be used to obtain exceptional situational awareness for assitive and autonomous driving systems. Although LiDARs demonstrate good performance in clean and clear weather conditions, their performance significantly deteriorates in adverse weather conditions such as those involving atmospheric precipitation. This may render perception capabilities of autonomous systems that use LiDAR data in learning based models to perform object detection and ranging ineffective. While efforts have been made to enhance the accuracy of these models, the extent of signal degradation under various weather conditions remains largely not quantified. In this study, we focus on the performance of an automotive grade LiDAR in snowy conditions in order to develop a physics-based model that examines failure modes of a LiDAR sensor. Specifically, we investigated how the LiDAR signal attenuates with different snowfall rates and how snow particles near the source serve as small but efficient reflectors. Utilizing our model, we transform data from clear conditions to simulate snowy scenarios, enabling a comparison of our synthetic data with actual snowy conditions. Furthermore, we employ this synthetic data, representative of different snowfall rates, to explore the impact on a pre-trained object detection model, assessing its performance under varying levels of snowfall.
\end{abstract}

\section{INTRODUCTION}

\begin{figure}[!h]
    \centering
    \includegraphics[scale= 0.5]{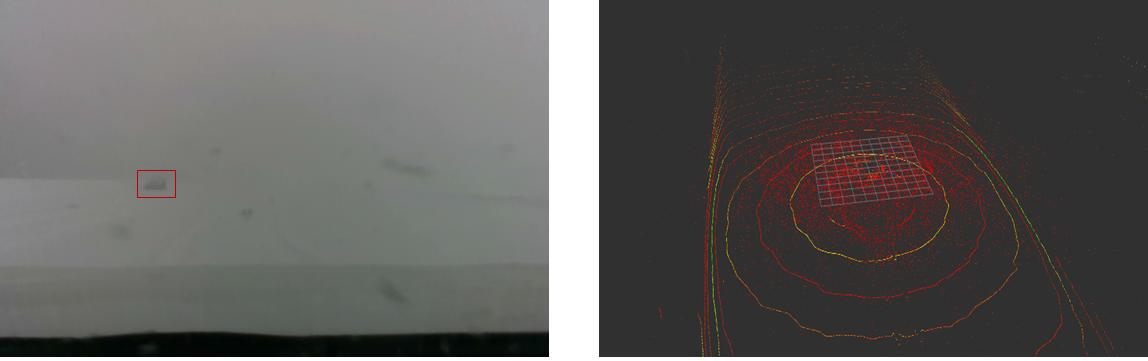}

    \caption{Left panel: RGB image of a heavy snowy situation, Right panel: the lidar point cloud representation}
    \label{fig:snowy_pic}
\end{figure}

Light Detection and Ranging (LiDAR) technologies are integral to Advanced Driver Assistance Systems (ADAS), providing essential perception capabilities involving object detection and recognition algorithms. Spatial and temporal resolution of modern LiDAR systems, especially under clean weather conditions, can be considered sufficient for highly automated driving systems. However, their effectiveness is considerably reduced in challenging weather scenarios such as fog, rain, or snow. As the transmitted light from a LiDAR travels over a distance, physical phenomena such as scattering and absorption of the energy result in exponential attenuation of the signal. During weather events, scattering caused by different atmospheric particles, such as water droplets and snowflakes, can result in even more pronounced attenuation. Additionally, the increased background signal in these conditions can lead to false detection. This effect is particularly acute near the sensor where the energy of the transmitted light is the greatest, and the backscattered signals from random particles can lead to misinterpretations by the detector, resulting in the potential oversight of actual targets amidst these distractions (refer to Fig \ref{fig:snowy_pic}). Hence, a common challenge in LiDAR-based perception systems in adverse weather is to distinguish false detection caused rain and snow particles from the actual objects. 

In order to address this challenge, de-noising filters have been proposed to enhance point cloud data. These filters are designed to remove backscattered light from snowflakes, which is identified as noise. The underlying principle of these methods is outlier detection algorithms. As an illustration, \cite{charron2018noising} introduced a dynamic radius outlier removal filter, tailored to the density variations in point clouds. This filter aims to eliminate noise from snow, which manifests as lower density areas in point clouds, while preserving environmental features characterized by higher density. While these de-noising filters effectively remove the noise from snow particles in an attempt to improve object detection quality, they do not compensate for signal attenuation caused by snow or rain in the atmosphere. Additionally, their reliance on point cloud density implies that in conditions of dense snowfall, the filters' accuracy will be reduced.

In addition to de-noising filters, various studies have focused on understanding how LiDAR performance is affected by adverse weather conditions and finding practical solutions to enhance object detection and recognition in such situations. These efforts involve both mathematical models that incorporate the physics of the atmosphere and LiDAR characteristics, and experimental validations. For instance, \cite{rasshofer2011influences} utilized Mie scattering theory to develop a mathematical model for fog and rainy conditions, assuming spherical water droplets. This model calculates the theoretical attenuation of LiDAR signals in foggy and rainy conditions. Additionally, it points out that backscattering from snowflakes or water droplets can result in false detection, resulting in false positives in object detection algorithms. Building upon this, \cite{hasirlioglu2016modeling} extended the model to heavy rain scenarios, where larger droplets and multiple scattering events become significant. They verified the accuracy of their model with experimental data from the CARISSMA rain simulator, comparing lidar signal degradation in actual experiments to their model predictions. Moreover, \cite{goodin2019predicting} investigated the correlation between rainfall intensity and visibility range, contrasting their findings with reference visibility ranges in clear weather conditions.

Indoor fog chambers are also used for investigating sensor performance and validating developed mathematical models against real conditions. For example, the CEREMA fog chamber, which allows adjustment of various weather parameters like fog particle size, meteorological visibility, and rain particle size and intensity, was utilized in \cite{bijelic2018benchmark, hasirlioglu2016test, hasirlioglu2017reproducible}. Notably, tests performed in fog chambers are always static, meaning neither the target nor the source is moving. Therefore, the generated data do not accurately represent moving cars or objects, and cannot be directly used in object detection algorithms.

In an effort to create LiDAR data representative of adverse weather conditions, \textit{Hahner et al., 2021} \cite{hahner2021fog} adapted real datasets captured in clear weather to simulate rainy and foggy conditions, utilizing the mathematical framework proposed by \cite{rasshofer2011influences}. They applied state-of-the-art detection methods to identify the most effective models for object detection in rainy and foggy conditions. In a similar vein, \cite{hahner2022lidar} modified LiDAR intensity data based on a simplified physical model of snowfall to create synthetic snowy scenarios. However, their model had limitations, such as treating snowflakes as ideal black objects that absorb all incident energy and considering the rest of the atmosphere as clear with no light attenuation. To enhance snow simulation, \cite{kilic2021lidar} treated snowflakes as spherical particles and applied Mie scattering theory to adjust point cloud intensity values from data obtained in clear conditions. They created physics-based synthetic data that took into account light attenuation due to backscattering by rain and snow particles.

While prior studies have shown the efficacy of training machine learning models using data simulated for adverse weather conditions, a direct comparison with real snowy weather data is lacking. Evaluating the accuracy of these simulations in representing real snowy conditions is therefore crucial. Such assessment is vital to ascertain the dependability of LiDAR-based object detection systems in genuine snowy environments. Concurrently, experimental studies have been carried out in snowy conditions. For example, \cite{jokela2019testing} conducted road tests in snowy weather, observing a decrease in LiDAR signal reception, as indicated by a reduced number of point clouds from targets. Additionally, \cite{kutila2020benchmarking} studied LiDAR performance in the harsh Arctic climate of Northern Europe, reporting signal attenuation due to turbulent snow and freezing temperatures. However, these investigations did not quantify the extent of signal degradation, nor did they explore its impact on the efficacy of object detection algorithms.

In this study, we aim to merge physical modeling of snowy conditions with empirical data. Initially, we develop a physics-based model that simulates the influence of snowy weather, specifically focusing on how snow particles in the atmosphere attenuate LiDAR signals. We quantify the resulting signal loss, as well as the effect of the back reflection of the light transmitted from LiDAR; specifically, closer to the emitter of the LiDAR. Our findings reveal that near the source, snow particles can substantially saturate the LiDAR system, leading to false detection or, in cases of intense snowfall, creating a barrier-like effect that impedes the system's perception capabilities. Furthermore, we have collected real LiDAR data in both clear and snowfall conditions. Using our model, we adjust the point cloud data obtained in clear conditions to mimic snowy scenarios. Comparing this synthetic data with actual point cloud data from snowy conditions allows us to more accurately gauge the impact snowfall imparts on LiDAR performance. Finally, using state-of-the-art machine learning algorithms, based on our synthetic data, we tested how the accuracy of object detection models degrades under different snowfall rates.

The structure of this paper is outlined as follows: Section~\ref{sec:method} dives into the physics underlying LiDAR attenuation across various environments. This section comprehensively covers the techniques used to calculate extinction and back-scattering coefficients in diverse conditions, ranging from clear and clean to rainy and snowy weather scenarios. Section \ref{sec:experment} details our experimental setup, including the specific conditions under which the experiments were carried out. In Section~\ref{sec:result}, we present the outcomes of modifying LiDAR point cloud data using our snow-focused physics-based model, and examine its impact on object detection algorithms. The paper concludes with Section~\ref{sec:discussion}, where we explore the implications of our methodology and outline directions for future work.

\section{Methodology} \label{sec:method}

\begin{figure}
    \centering
    \includegraphics[scale = 0.33]{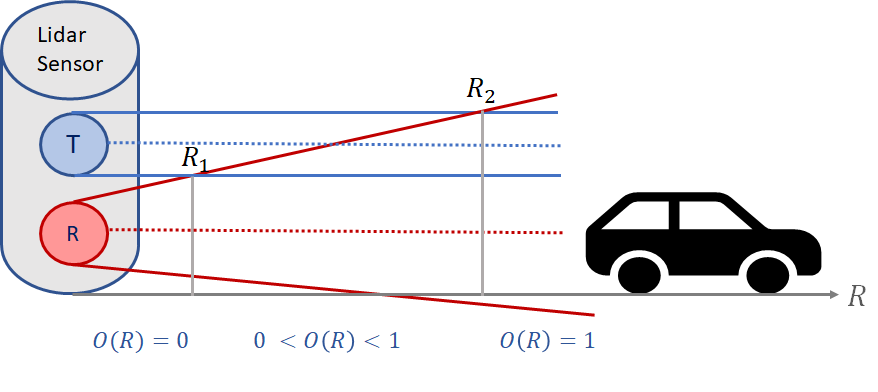}
    \caption{Schematic representation of a LiDAR system}
    \label{fig:lidar_schamtics}
\end{figure}

LiDAR systems determine target distances by emitting light pulses and measuring the time required for the back-scattered light to return (Fig~\ref{fig:lidar_schamtics}). As a laser beam traverses through the atmosphere, it encounters particulates that cause attenuation. The intensity of the light backscattered from a target at distance $R$, denoted as $P(R)$, is given by:

\begin{equation}
    P(R) = C \beta_{0} O(R) \frac{e^{-2\int_{0} ^{R} \alpha(R) dR}}{R^2},
    \label{equ:lidar_equ}
\end{equation}

In (\ref{equ:lidar_equ}), $C$ represents a constant characterizing the LiDAR system, which includes factors such as the area of the receiving mirror and the system's efficiency. The term $\beta_{0}$ refers to the reflectivity of the target. The overlap function $O(R)$ for biaxial LiDAR systems (Fig~\ref{fig:lidar_schamtics}) is defined as the ratio of the area illuminated by the transmitter ($A_T$) to the area illuminated by the receiver ($A_R$). This function, as detailed in \cite{hahner2022lidar}, is expressed as:

\begin{equation}
        O(R)= 
\begin{cases}
    0,&  R\leq R_{1}\\
    \frac{R-R_1}{R_2 - R_1},&  R_1 \leq R\leq R_{2}\\
    1,& R\leq R_{2}
\end{cases}
\end{equation}

Here, $R_1$ is the minimum effective range of the LiDAR (outside of which it cannot detect objects), typically between 0.5\,m to 1\,m for ADS LiDARs \cite{veldoyne} as can be found in the specification of each LiDAR system. $R_2$ is the distance at which the overlap between $A_T$ and $A_R$ becomes complete.

\begin{figure}
    \centering
    \includegraphics[scale = 0.40]{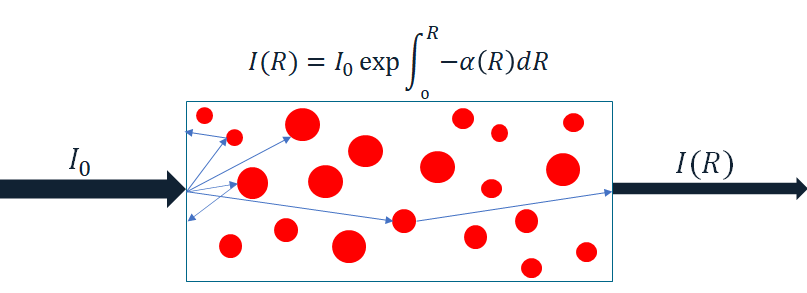}
    \caption{Light attenuation in a medium}
    \label{fig:attenuation_schema}
\end{figure}

The extinction coefficient $\alpha$ represents the likelihood per unit length that a photon will be scattered or absorbed, thus being removed from the beam. This coefficient is higher in mediums with more particles, leading to greater attenuation of the laser beam (Fig~\ref{fig:attenuation_schema}). The total one-way transmission loss in a medium is:

\begin{equation}
    T(R) = e^{-\int_{0} ^{R} \alpha(R) dR}.
    \label{eq:transmission}
\end{equation}

First-principles physics-based approaches have proven highly effective  for LiDAR data analysis in atmospheric science applications, where the scattering properties and vertical density profiles of various atmospheric 
constituents have been extensively characterized~\cite{farhani2019improved, farhani2023bayesian, farhani2019optimal}. These methods not only enable accurate retrieval of atmospheric profiles but also provide rigorous uncertainty quantification of the retrieved quantities. 

In this work, we extend these physics-based principles to automotive-grade LiDAR systems, analyzing how laser light interacts with precipitation particles including snow, rain, and fog, and quantifying the resulting signal attenuation that impacts 3D object detection performance.

\subsection{Light Attenuation in different Mediums}
In clean and clear weather conditions, with laser wavelengths in the visible and infrared spectrum, atmospheric molecules are substantially smaller than the laser wavelength. Therefore, the extinction coefficient of air molecules (\(\alpha\)) in the medium can be calculated using Rayleigh scattering theory. For instance, at a laser wavelength of 900\,nm, \(\alpha\) is approximately \(1.52 \times 10^{-6}\) m\(^{-1}\) \cite{heinz2007rayleigh}. LiDAR manuals typically provide the maximum visibility range of the system, which is often in the range of 100\,m to 200\,m. Consequently, the transmission defined in (\ref{eq:transmission}) at \(R_{max} = 100\) m is close to one, indicating that light attenuation in clean and clear conditions for ADS LiDAR systems is not significant. Additionally, the particle sizes compared to the wavelength are too small to consider their backscattering characteristics. In contrast, during rain, foggy and snowy conditions, water droplet sizes are comparable to the wavelength, leading to more significant light attenuation. Here, the light attenuation for rain and fog, as well as in snowy condition will be discussed.  

\subsubsection{Light Attenuation in Rain and Foggy Conditions}
Water droplets are relatively small, however they are efficient reflectors and can generate false positives within a range closer to the source (\(R \leq 10\) m). By employing Mie scattering theory for spherical particles, we can calculate the extinction efficiency (\(Q_{ext}(D)\)) and backscattering efficiency (\(Q_{b}(D)\)). Thus, the extinction coefficient (\(\alpha\)) and backscattering coefficient (\(\beta\)) of rain and fog particles can be estimated as:

\begin{equation}
\begin{aligned}
    \alpha &= \int_{D=0} ^{D=\infty} D^2 N(D) Q_{ext}(D) \, dD; \\
    \beta &= \int_{D=0} ^{D=\infty} D^2 N(D) Q_{b}(D) \, dD,
\end{aligned}
\label{equ:alpha_beta}
\end{equation}

where \(D\) is the diameter of raindrops, and \(N(D)\) is the distribution of droplet diameters for rain and fog.

\subsubsection{Light Attenuation in Snowfall Conditions}
The analysis of light attenuation and backscattering by snowflakes is complex due to the fact that typical snow crystal are of non-spherical shape. However, for forward diffraction purposes, an aggregate of spheres, whose cross-sectional areas are equivalent to those of the irregular snow particles, is a reasonable approximation. Consequently, it is customary to employ the spherical assumption for snow particles larger than \(1 \mu m\) \cite{hodkinson1963light, hahner2022lidar}. Furthermore, since most snow particles are larger than near-infrared (NIR) wavelengths, the geometrical optics approximation may be more appropriate than Mie theory. This suggests that the extinction of light in falling snow might be wavelength-independent. Contrarily, numerous studies, such as those by \cite{sola1978multispectral, chen2003frequency, tsang2021global}, have established that snow particle-induced light extinction does depend on wavelength. To accurately calculate extinction efficiency with wavelength considerations, \cite{sola1978multispectral} proposed a method to compute the relative diffracted intensity of electromagnetic energy at a wavelength \(\lambda\) when diffracted by a sphere with radius r, leading to an approximation of \(Q_{eff}\) as:

\begin{equation}
    Q_{eff} = \exp(-0.88 \kappa) + 1.
    \label{equ:Q_eff}
\end{equation}

In this context, \(\kappa = \frac{2 \pi r r_d}{\lambda L}\), where \(r\) represents the radius of the snow particle, \(r_d\) is the radius of the detector, and \(L\) denotes the distance between the particle and the detector. As a result, akin to the estimations for fog and rain, knowing \(N(D)\), (\ref{equ:alpha_beta}) can be utilized to approximate \(\alpha\) under snowy conditions. 

Empirical research has been undertaken to explore the distribution of snow particles across various diameters. Drawing on these investigations, a mathematical model has been suggested \cite{best1950size, zhang2013review}:
\begin{equation}
    N(D) = N_0 \exp(-\Lambda D),
    \label{equ:snow_dist}
\end{equation}
where \(N_0\) and \(\Lambda\) are parameters used for fitting the model. Several experiments have endeavored to determine general values for \(N_0\) and \(\Lambda\), and to correlate these values with the snow rate (\(Sr\)) \cite{zhang2013review}.

Finally the reflectance of the snow grains at a specific wavelength, depends on the size of the grain, using experimental data from \cite{singh2010hyperspectral}, the averaged value of $\beta = 0.4$ can be used.

\subsection{Simulating the Effect of Light Attenuation under Snowfall Condition}
With different \(\alpha\) values representing various snowfall rates, the power loss at a distance \(R\) from the source can be estimated, which indicates the intensity loss in the point cloud. By integrating (\ref{equ:snow_dist}) to determine \(N(D)\) and then applying it to (\ref{equ:alpha_beta}), we can derive \(\alpha\) for a range of snowfall rates (\(Sr\)). Accurately quantifying the laser power loss at varying snowfall rates is essential for understanding its effect on point cloud data analysis. Therefore, employing (\ref{equ:lidar_equ}), we formulate the ratio of received signal powers at a distance \(R\) from a target, characterized by reflectivity \(\beta_0\), in snowy conditions compared to clear weather conditions. This relationship is mathematically expressed as:
\begin{equation}
    \frac{P_{snow}(R)}{P_{clear}(R)} = \frac{\exp(-2 \alpha_{snow} R)}{\exp(-2 \alpha_{clear} R)},
    \label{equ:power_loss_snow}
\end{equation}
Equipped with this equation and the known extinction coefficient for snow, it becomes feasible to extrapolate from the intensity of the backscattered signal of an object in clear and clean conditions to its intensity under snowy conditions. This approach bridges the gap in understanding how snow impacts LiDAR measurements, enhancing our capability to model and analyze environmental data under varying weather scenarios.

\subsection{Object Detection Algorithms for Point Cloud Data} \label{Odetection}

For our study, the LiDAR point cloud semantic segmentation model SphereFormer was used \cite{lai2023spherical}. This contribution is currently the state-of-the-art model for LiDAR semantic segmentation as it ranks 1st on both NuScenes\cite{Nuscenes} and SemanticKITTI\cite{behley2019iccv} semantic segmentation benchmarks with 81.9\% and 74.8\% respectively. This deep learning model utilizes a Sparse Convolution Network with radial window attention. This feature significantly improves the accuracy of distant points from 13.3\% to 30.4\%. The model trained on SemanticKITTI data is used for this study.
\section{Experimental Setup} \label{sec:experment}

\subsection{Data Acquisition System}
\begin{figure}
    \centering
 
    \includegraphics[scale = 0.3]{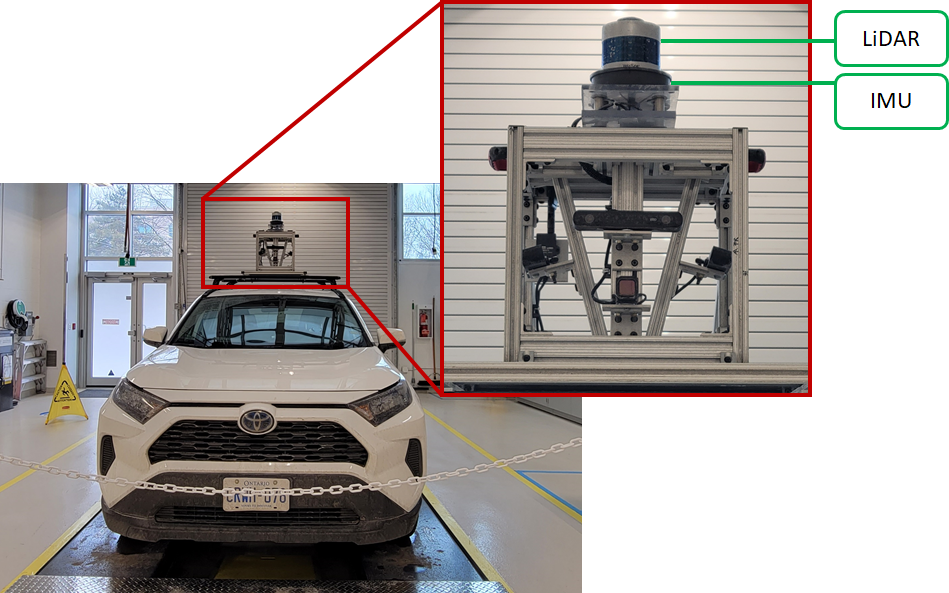}
    \includegraphics[scale = 0.29]{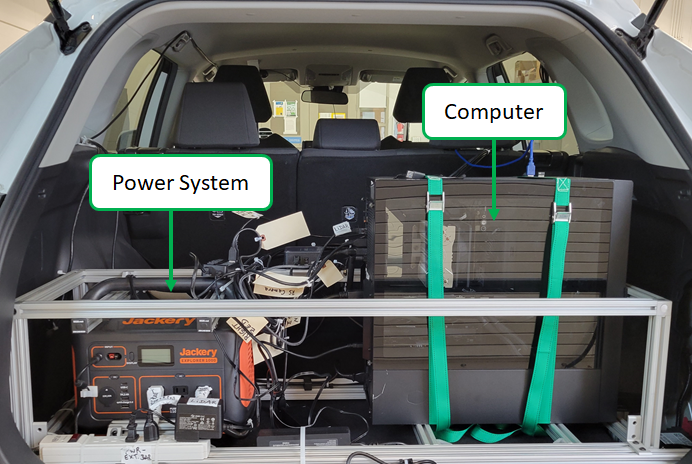}
    \caption{Top Panel: Instrument installed at the roof of the vehicle. Bottom Panel: The computing platform and the mobile power source.}
    \label{fig:jetta}
\end{figure}

The vehicle used for data acquisition was a 2021 RAV4 equipped with a data acquisition system encompassed of sensors, a computing platform to host the acquisition system software, and auxiliary components that were a mobile power source, and cables used for power and data transportation. While the focus of this study was on understanding the LiDAR signal degradation in the snowy conditions, additional to a LiDAR instrument, infrared camera, RGB camera, radar, IMU, and GNSS-RTK sensors were installed on the roof of the vehicle (Figure~\ref{fig:jetta}, top panel) to facilitate future perception and localization research for autonomous vehicles. Moreover, the computing platform and the mobile power source were installed in the trunk (Figure~\ref{fig:jetta}, bottom panel). Specification of the sensors are provided in Table~\ref{tab:sensorspecs}. 

The software platform of the data acquisition is based on Robot Operating System (ROS) \cite{ros} which is an open-source software contribution and is widely used in robotics research. Each of the sensors was integrated into the ROS ecosystem through a driver \textit{node}, which is programmed to parse binary data transmitted by the sensors in order to \textit{publish} them under a \textit{topic} as \textit{messages}. The  published messages are recorded in a \textit{bag} file. A GUI (graphical user front-end) was developed to enable an operator to interact and record sensor data into bag files.

The experiments that were used in this study carried out on three different days, as explained in Table~\ref{tab:condition}. Two dates had snowy conditions and they were used to simulate the snow particles distribution, and one date corresponded to clean and clear conditions and the point cloud data were repurposed to simulated snowy condition.

\begin{table}[]
\centering
\begin{tabular}{|ll|}
\hline
\multicolumn{2}{|c|}{\cellcolor[HTML]{ECF4FF} \textbf{Specifications}}                                                                                                                                                                              \\ \hline
\multicolumn{1}{|l|}{Vehicle}                   & 2021 RAV4                                                                                                                                                                               \\ \hline
\multicolumn{1}{|l|}{LiDAR (1)}                     & \begin{tabular}[c]{@{}l@{}}Velodyne, VLP-32\\  \( 360^\circ \) horizontal FOV \( 40^\circ \) vertical FOV\\  200 m max range, 32 channels\end{tabular}                                                          \\ \hline
\multicolumn{1}{|l|}{Infrared camera (1)}           & \begin{tabular}[c]{@{}l@{}}Intel RealSense D415, 850 \,nm NIR, \\ 1920 $\times$ 1080 active pixels\\  \( 69.4^\circ \) horizontal FOV, \( 42.5^\circ \) vertical FOV\end{tabular}                                  \\ \hline
\multicolumn{1}{|l|}{RGB camera (1)}                 & \begin{tabular}[c]{@{}l@{}}Intel RealSense D415, \\ 1920 $\times$ 1080 active pixels\\  \( 69.4^\circ \) horizontal FOV, \( 42.5^\circ \) vertical FOV\end{tabular}                                  \\ \hline
\multicolumn{1}{|l|}{GNS-RTK (1)}                   & \begin{tabular}[c]{@{}l@{}}Swift NAV Piksi Multi,\\ GPS, GLONASS, Galileio, BeiDou \\ constellations, RTK relative accuracy: \\ $\sim$1cm horizontal, $\sim$1.5cm vertical\end{tabular} \\ \hline
\multicolumn{1}{|l|}{Radar Array (3)}               & \begin{tabular}[c]{@{}l@{}}GNSSTexas Instruments mm Wave\\ 76-81GHz\\ Single plane sensing w/$\pm$ \( 60^\circ \) horizontal FOV\end{tabular}                                                        \\ \hline
\multicolumn{1}{|l|}{IMU (1)}                       & \begin{tabular}[c]{@{}l@{}}Xsens 680G INS\\  Sensor fusion performance:\\  Roll/pitch: 0.2 RMS\\  Yaw/heading: \( 0.5^\circ \) RMS\\ Position: 1cm CEP, velocity: 0.05m/s RMS\end{tabular}          \\ \hline
\multicolumn{1}{|l|}{Instrument Cluster Camera} & 720p consumer grade webcam                                                                                                                                                                    \\ \hline
\multicolumn{1}{|l|}{CAN-bus monitor (1)}           & OBD-II CAN-bus monitor w/ SocketCan support                                                                                                                                             \\ \hline
\multicolumn{1}{|l|}{In-vehicle power (1)}          & 1000Wh rechargeable power station                                                                                                                                                       \\ \hline

\end{tabular}
\caption{Specification of sensors used during measurements.}
\label{tab:sensorspecs}
\end{table}

\begin{table}[]
\centering
\begin{tabular}{|l|l|l|}
\hline
\rowcolor[HTML]{ECF4FF} 
Date        & Speed & Weather Condition                                                                                               \\ \hline
18 Feb 2022 & 60kph & \begin{tabular}[c]{@{}l@{}}Day, moderate snowfall, limited visibility \\ snow covered road surface\end{tabular} \\ \hline
25 Feb 2022 & 60kph & \begin{tabular}[c]{@{}l@{}}Day, heavy snowfall, limited visibility \\  snow covered road surface\end{tabular}   \\ \hline
22 May 2022 & 60kph & \begin{tabular}[c]{@{}l@{}}Sunny, excellent visibility\\  dry road surface\end{tabular}                         \\ \hline
\end{tabular}
\caption{Specifications of the dates of experiments.}
\label{tab:condition}
\end{table}

\section{Results}\label{sec:result}
In this section, using the methodology introduced in \ref{sec:method}, we compute the extinction coefficient of snow under different falling rates, and show how different snow rates can effect the signal degradation. Then we modify the point cloud data from clean and clear conditions to simulate the snowy conditions. And, we employ the trained object detection algorithm (introduced in Section~\ref{Odetection}) to investigate how snow conditions can degrade the accuracy of the object detection model.  

\subsection{Calculating Extinction Coefficient}

\begin{figure}
    \centering
    \includegraphics[scale=0.25]{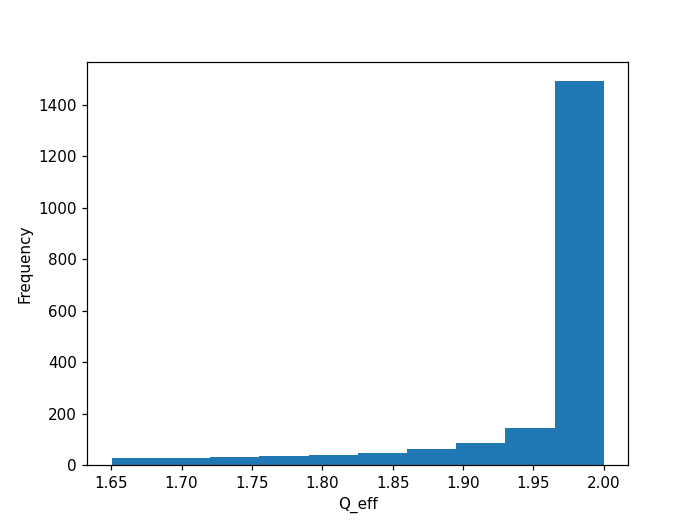}
    \caption{Variation of $Q_{eff}$ across different sizes of snow particles.}
    \label{fig:Q_eff}
\end{figure}

In our analysis, we compute $Q_{eff}$ utilizing a diffraction methodology grounded in geometrical optics. The derived values of $Q_{eff}$ are influenced by the dimensions of the snow particles and their respective distances from the light source. Following (\ref{equ:Q_eff}), we created a distribution of $Q_{eff}$ values for a set of 2000 snow particles. These particles, with sizes spanning from 0.05\,mm to 5\,mm, were defined according to (\ref{equ:snow_dist}). Subsequently, these particles were allocated randomly across distances ranging from 0\,m to 50\,m, in line with the typical operating range of the LiDAR used in this study. The median $Q_{eff}$ value from this analysis, shown in Figure~\ref{fig:Q_eff}, was established as 1.97.

\begin{figure}
    \centering
    \includegraphics[scale = 0.25]{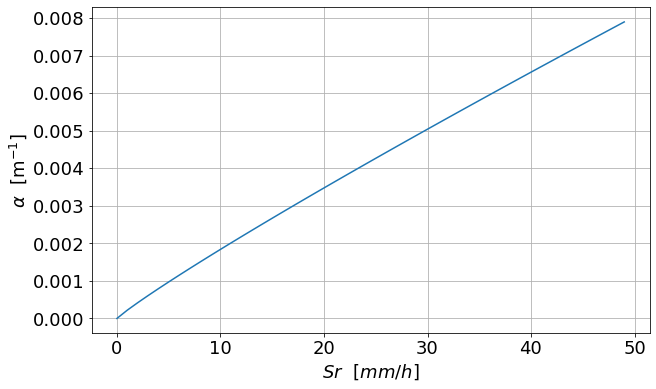}
    \includegraphics[scale = 0.16]{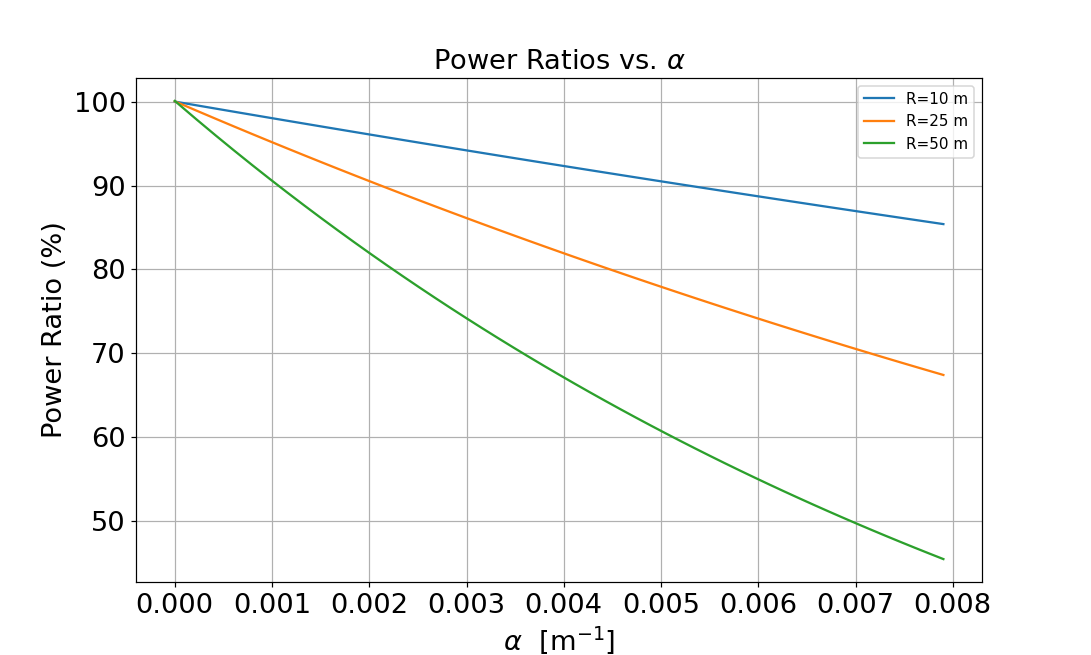}

    \caption{High rate of snowfall, corresponds to larger $\alpha$.}
    \label{fig:alpha}
\end{figure}

After determining $Q_{eff}$, the extinction coefficient $\alpha$ was calculated using (\ref{equ:alpha_beta}). As illustrated in Fig~\ref{fig:alpha} (left panel), there is an observable correlation between various snowfall rates (\(Sr\)) and $\alpha$. This variation in \(\alpha\) with distinct \(Sr\) values facilitates the establishment of a relationship between \(Sr\) and the power loss ratio, as depicted in Fig~\ref{fig:alpha} (right panel). Therefore, applying (\ref{equ:power_loss_snow}) for a specific distance \(R\), enables precise quantification of laser intensity loss under different snowfall rates.

\subsection{Point Cloud Simulation} 
\begin{figure}
    \centering
    \includegraphics[scale=0.45]{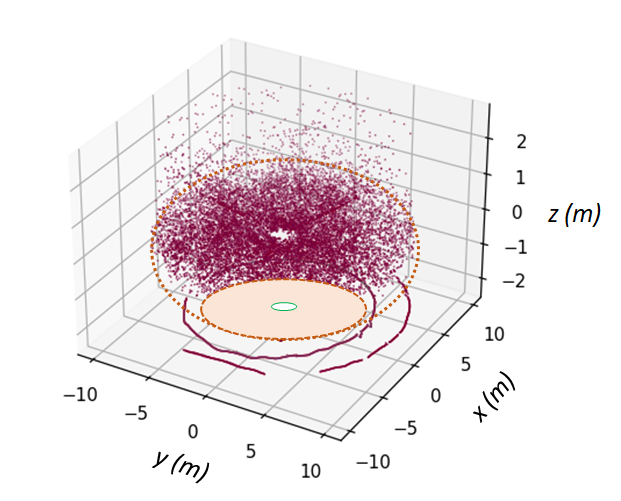}
    \caption{Visualization of the snow distribution on 25 February 2022.}
    \label{fig:snow_bulk}
\end{figure}

Point cloud data from clear weather conditions can be adapted to simulate snowy environments with varying snowfall rates. Creating realistic synthetic data necessitates accounting for the presence of snow particles near the LiDAR source, as these particles can result in false positives. To comprehend the distribution of these snow particles, we examine the point cloud visualized on 25 February 2022 (Fig~\ref{fig:snow_bulk}). The backscatter from snow particles predominantly occurs within a 10-meter radius of the source, forming a hemispherical shape. Notably, within a 0.5-meter radius, a void is observed due to the lack of data, creating a central gap in the hemispherical formation. To fabricate synthetic snow particles, it is essential to calculate the total number of particles expected within the desired volume. This calculation is based on integrating (\ref{equ:snow_dist}) across the particle size spectrum for different snowfall rates (\(Sr\)), we assumed that the boundary of integration is from 0.001 mm to 15 mm.
\begin{equation}
    N_{total} = \int_{0.001} ^{15} N_0 \exp(-\Lambda D) \, dD.
    \label{equ:snow_total}
\end{equation}

For initial values of \(N_0\) and \(\Lambda\), we began with \(N_0 = 0.8 \, \text{cm}^{-4}\) and \(\Lambda = 0.41 \, \text{(Sr)}^{-0.21}\, \text{cm}^{-1}\), as per \cite{zhang2013review}, with \(Sr\) representing the snowfall rate in mm/h. Upon analyzing actual snowfall data from 25 February 2022 and 18 February 2022, we adjusted these parameters to \(N_0 = 0.5 \, \text{cm}^{-4}\) and \(\Lambda = 0.41 \, \text{(Sr)}^{-0.31}\, \text{cm}^{-1}\). These revised values enable the estimation of the total snow volume detectable by LiDAR for varying \(Sr\) rates. To simulate a realistic snow particle distribution, we utilized coordinates from detected snow particles on 25 February 2022, adding Gaussian noise to these coordinates and selecting them randomly for our simulations. The choice of 25 February 2022, a day with heavy snowfall at approximately 35 mm/h, was strategic for these purposes.

\begin{figure}
    \centering
    \includegraphics[scale = 0.35]{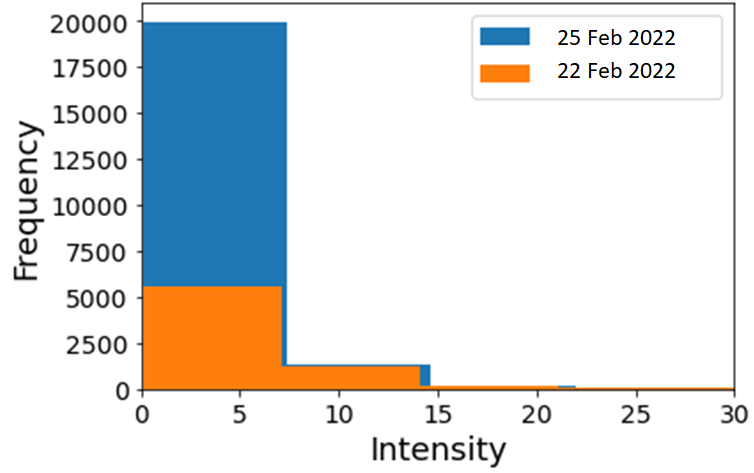}
    \caption{Histogram of the intensity of back scattered snow particles, randomly selected from lidar scans on 18 Feb 2022 and 25 Feb 2022.}
    \label{fig:hist_intensity}
\end{figure}

To ascertain the intensity of snow particles in our simulations, we examined a random assortment of LiDAR point cloud scans from 25 February 2022 and 18 February 2022. Histograms from two such scans were analyzed. The intensity of the backscattered signals from snow particles predominantly fell below 5, with an average hovering around 1.5, as illustrated in Fig~\ref{fig:hist_intensity}. Utilizing this average intensity, and referencing (\ref{equ:lidar_equ}), enabled us to calculate the LiDAR constant (C). With the C value established and an average \(\beta\) of 0.4 applied across all particles, we could then determine the intensity of the backscattered signals for the specifically chosen coordinates, which served as proxies for snow particles.

\subsubsection{Physics-Based Simulated Point Clouds for Snowy Conditions}
\begin{figure*}
\centering
    \includegraphics[scale = 0.51]{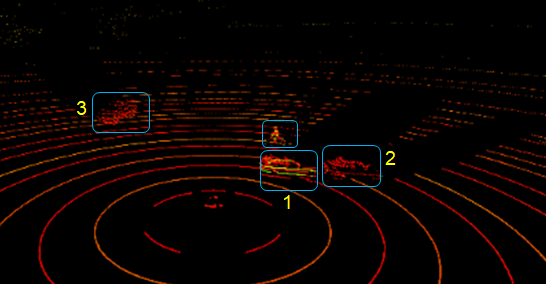}
    \includegraphics[scale = 0.4]{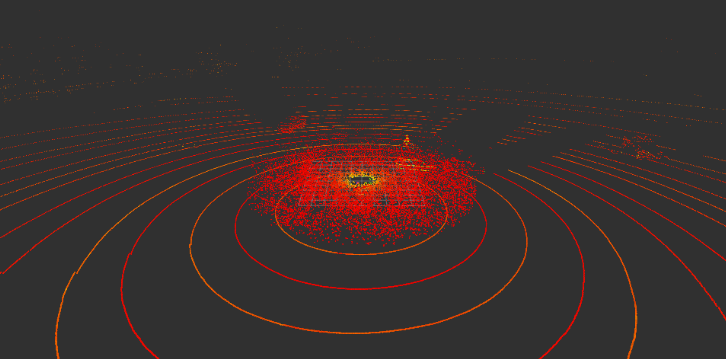}
    \caption{Left panel: Visualization of point cloud data from 22 May 2022 LidDAR measurements. Right Panel: Visualization of point cloud modified based on using our physics-based model for $Sr = 5$ mm/h.}
    \label{fig:clean_Scan}
\end{figure*}

We present the outcomes of applying our physics-based model to adapt point cloud data from clear weather conditions. We selected LiDAR measurements from 12 May 2022 as a baseline for clean weather conditions, as shown in Fig~\ref{fig:clean_Scan} (left panel). In the point cloud visualization, two cars and a cyclist are discernible at a distance of 10 meters from the LiDAR source, while another car is positioned at a farther distance of 22 meters. Utilizing our model, we altered this scan to demonstrate the impact of snow under different snowfall rates (\(Sr\)). At an \(Sr\) of 5 mm/h (see Fig~\ref{fig:clean_Scan}, left panel), snow particles are clearly present and significantly blur the visibility of car number 2. Additionally, due to light attenuation, car 3 becomes less distinct. It also becomes challenging to differentiate between the cyclist, car number 1, and the cyclist.


\subsubsection{Object Detection Under Snowy Condition}
\begin{figure*}
\centering
    \includegraphics[scale = 0.45]{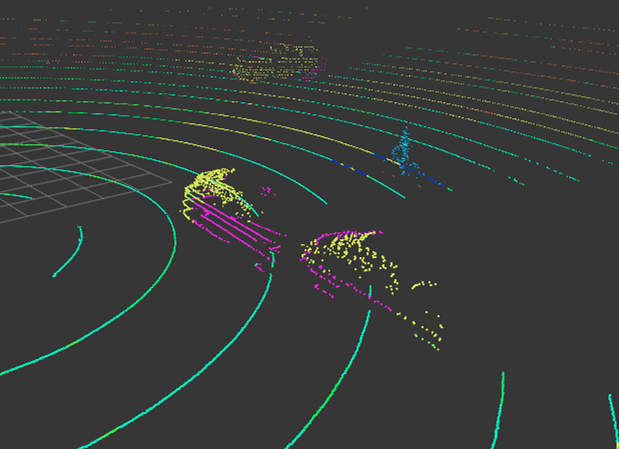}
    \includegraphics[scale = 0.45]{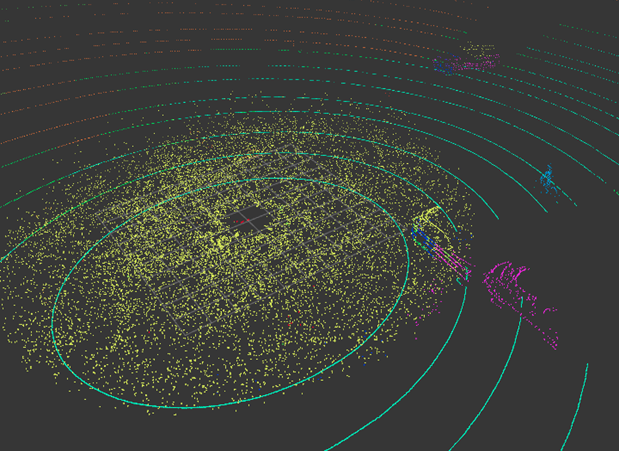}
\caption{Left Panel: objects detected under clean condition. Right Panel: Objects detected under snow condition. }
    \label{fig:Odetection_clean}
\end{figure*}

We utilized a pre-trained model that described in Section~\ref{Odetection}, for the object detection task. This model effectively identifies the cyclist, as shown in the left panel of Fig~\ref{fig:Odetection_clean}. However, its performance in car detection demonstrates some limitations. Specifically, certain portions of the cars are inaccurately categorized as ``unknown'' (indicated by a purple color), highlighting the model's partial non-generalizability. This is likely due to the model being trained on a different dataset, leading to a degree of uncertainty in car detection.

Additionally, the model's performance is significantly challenged in simulated snowy conditions, as seen in the right panel of Fig~\ref{fig:Odetection_clean}. It experiences difficulty in distinguishing between snow particles and a car, a problem primarily caused by the high reflectivity of nearby snow. This reflectance confuses the model and hampers its ability to accurately separate the car from surrounding snow particles. Our synthetic data indicate that within the first 10 meters from the source, the system registers a high number of false positives, making it exceedingly difficult to differentiate between objects and snow. While the signal does experience attenuation, object detection is still feasible. However, the initial 10-meter zone emerges as the most problematic area. These findings are consistent with those reported in previous LiDAR studies, such as those documented in \cite{michaud2015towards}.

\section{Discussion and Future Direction} \label{sec:discussion}

In this study, we developed a physics-based model designed to simulate the impact of LiDAR degradation in snowy weather conditions. A key strength of our approach is the use of actual point cloud data gathered during snowfall. This data was instrumental in accurately mapping snow distribution near the LiDAR system, allowing us to generate realistic synthetic data. Our focus was on understanding how backscattering signal from snow particles contributes to false positives in detection, and this was informed by our direct measurements of snow conditions.

Additionally, we applied the diffraction theory to calculate the extinction efficiency and coefficient of snow particles. With these calculations, we could adapt point cloud data from clear weather conditions to represent snowy environments. We also investigated the impact of light degradation at different rates of snowfall.  

However, our work has a notable limitation: the clean and snowy datasets were not captured from the same scene, which poses a challenge in accurately evaluating our model's effectiveness. Our future research aims to address this by collecting data from identical scenes under varying weather conditions. Furthermore, we plan to determine the LiDAR constant $C$ by using a known reflective target at a fixed distance, which will refine our physics-based model and reduce reliance on the simplifications currently in place.

The object detection algorithm also demonstrated deficiencies in identifying objects from our dataset, underscoring the need for model retraining to better suit our specific data. One main reason for the shortcoming of the object detection model is that our data is acquired from a 32 channel Velodyne LiDAR, however, the pre-train model used for the object detection task is based on KITTI dataset which is a 64 channel Velodyne LiDAR system. We intend to leverage the U-Net encoder-decoder structure of the mentioned model, focusing on the features extracted from the final decoder layers. These will be used to train a shallow CNN classifier with labeled data from our target domain, employing a transfer learning approach to enhance the model's accuracy in our specific context.

\bibliographystyle{unsrt}  

\bibliography{ADAS.bib}

\begin{thebibliography}{10}

\bibitem{charron2018noising}
Nicholas Charron, Stephen Phillips, and Steven~L Waslander.
\newblock De-noising of lidar point clouds corrupted by snowfall.
\newblock In {\em 2018 15th Conference on Computer and Robot Vision (CRV)}, pages 254--261. IEEE, 2018.

\bibitem{rasshofer2011influences}
Ralph~H Rasshofer, Martin Spies, and Hans Spies.
\newblock Influences of weather phenomena on automotive laser radar systems.
\newblock {\em Advances in radio science}, 9:49--60, 2011.

\bibitem{hasirlioglu2016modeling}
Sinan Hasirlioglu, Igor Doric, Christian Lauerer, and Thomas Brandmeier.
\newblock Modeling and simulation of rain for the test of automotive sensor systems.
\newblock In {\em 2016 IEEE Intelligent Vehicles Symposium (IV)}, pages 286--291. IEEE, 2016.

\bibitem{goodin2019predicting}
Christopher Goodin, Daniel Carruth, Matthew Doude, and Christopher Hudson.
\newblock Predicting the influence of rain on lidar in adas.
\newblock {\em Electronics}, 8(1):89, 2019.

\bibitem{bijelic2018benchmark}
Mario Bijelic, Tobias Gruber, and Werner Ritter.
\newblock A benchmark for lidar sensors in fog: Is detection breaking down?
\newblock In {\em 2018 IEEE Intelligent Vehicles Symposium (IV)}, pages 760--767. IEEE, 2018.

\bibitem{hasirlioglu2016test}
Sinan Hasirlioglu, Alexander Kamann, Igor Doric, and Thomas Brandmeier.
\newblock Test methodology for rain influence on automotive surround sensors.
\newblock In {\em 2016 IEEE 19th International Conference on Intelligent Transportation Systems (ITSC)}, pages 2242--2247. IEEE, 2016.

\bibitem{hasirlioglu2017reproducible}
Sinan Hasirlioglu, Igor Doric, Alexander Kamann, and Andreas Riener.
\newblock Reproducible fog simulation for testing automotive surround sensors.
\newblock In {\em 2017 IEEE 85th Vehicular Technology Conference (VTC Spring)}, pages 1--7. IEEE, 2017.

\bibitem{hahner2021fog}
Martin Hahner, Christos Sakaridis, Dengxin Dai, and Luc Van~Gool.
\newblock Fog simulation on real lidar point clouds for 3d object detection in adverse weather.
\newblock In {\em Proceedings of the IEEE/CVF International Conference on Computer Vision}, pages 15283--15292, 2021.

\bibitem{hahner2022lidar}
Martin Hahner, Christos Sakaridis, Mario Bijelic, Felix Heide, Fisher Yu, Dengxin Dai, and Luc Van~Gool.
\newblock Lidar snowfall simulation for robust 3d object detection.
\newblock In {\em Proceedings of the IEEE/CVF Conference on Computer Vision and Pattern Recognition}, pages 16364--16374, 2022.

\bibitem{kilic2021lidar}
Velat Kilic, Deepti Hegde, Vishwanath Sindagi, A~Brinton Cooper, Mark~A Foster, and Vishal~M Patel.
\newblock Lidar light scattering augmentation (lisa): Physics-based simulation of adverse weather conditions for 3d object detection.
\newblock {\em arXiv preprint arXiv:2107.07004}, 2021.

\bibitem{jokela2019testing}
Maria Jokela, Matti Kutila, and Pasi Pyyk{\"o}nen.
\newblock Testing and validation of automotive point-cloud sensors in adverse weather conditions.
\newblock {\em Applied Sciences}, 9(11):2341, 2019.

\bibitem{kutila2020benchmarking}
Matti Kutila, Pasi Pyyk{\"o}nen, Maria Jokela, Tobias Gruber, Mario Bijelic, and Werner Ritter.
\newblock Benchmarking automotive lidar performance in arctic conditions.
\newblock In {\em 2020 IEEE 23rd International Conference on Intelligent Transportation Systems (ITSC)}, pages 1--8. IEEE, 2020.

\bibitem{veldoyne}
{\em {Robotics Specifications Veldoyne}}, 2018.

\bibitem{farhani2019improved}
Ghazal Farhani, Robert~J Sica, Sophie Godin-Beekmann, G{\`e}rard Ancellet, and Alexander Haefele.
\newblock Improved ozone dial retrievals in the upper troposphere and lower stratosphere using an optimal estimation method.
\newblock {\em Applied optics}, 58(6):1374--1385, 2019.

\bibitem{farhani2023bayesian}
Ghazal Farhani, Giovanni Martucci, Tyler Roberts, Alexander Haefele, and Robert~J Sica.
\newblock A bayesian neural network approach for tropospheric temperature retrievals from a lidar instrument.
\newblock {\em International Journal of Remote Sensing}, 44(5):1611--1627, 2023.

\bibitem{farhani2019optimal}
Ghazal Farhani, Robert~J Sica, Sophie Godin-Beekmann, and Alexander Haefele.
\newblock Optimal estimation method retrievals of stratospheric ozone profiles from a dial.
\newblock {\em Atmospheric Measurement Techniques}, 12(4):2097--2111, 2019.

\bibitem{heinz2007rayleigh}
Tony~F Heinz.
\newblock Rayleigh scattering spectroscopy.
\newblock In {\em Carbon Nanotubes: Advanced Topics in the Synthesis, Structure, Properties and Applications}, pages 353--369. Springer, 2007.

\bibitem{hodkinson1963light}
J~Raymond Hodkinson.
\newblock Light scattering and extinction by irregular particles larger than the wavelength.
\newblock {\em ICES Electromagnetic scattering}, 63:87, 1963.

\bibitem{sola1978multispectral}
Marcos~C Sola and Richard~J Bergemann.
\newblock Multispectral propagation measurements through snow (a).
\newblock {\em Journal of the Optical Society of America (1917-1983)}, 68:541, 1978.

\bibitem{chen2003frequency}
Chi-Te Chen, Leung Tsang, Jianjun Guo, Alfred~TC Chang, and Kung-Hau Ding.
\newblock Frequency dependence of scattering and extinction of dense media based on three-dimensional simulations of maxwell's equations with applications to snow.
\newblock {\em IEEE transactions on geoscience and remote sensing}, 41(8):1844--1852, 2003.

\bibitem{tsang2021global}
Leung Tsang, Michael Durand, Chris Derksen, Ana~P Barros, Do-Hyuk Kang, Hans Lievens, Hans-Peter Marshall, Jiyue Zhu, Joel Johnson, Joshua King, et~al.
\newblock Global monitoring of snow water equivalent using high frequency radar remote sensing.
\newblock {\em The Cryosphere Discussions}, 2021:1--57, 2021.

\bibitem{best1950size}
AC~Best.
\newblock The size distribution of raindrops.
\newblock {\em Quarterly journal of the royal meteorological society}, 76(327):16--36, 1950.

\bibitem{zhang2013review}
L~Zhang, X~Wang, MD~Moran, and J~Feng.
\newblock Review and uncertainty assessment of size-resolved scavenging coefficient formulations for below-cloud snow scavenging of atmospheric aerosols.
\newblock {\em Atmospheric Chemistry and Physics}, 13(19):10005--10025, 2013.

\bibitem{singh2010hyperspectral}
SK~Singh, AV~Kulkarni, and BS~Chaudhary.
\newblock Hyperspectral analysis of snow reflectance to understand the effects of contamination and grain size.
\newblock {\em Annals of Glaciology}, 51(54):83--88, 2010.

\bibitem{lai2023spherical}
Xin Lai, Yukang Chen, Fanbin Lu, Jianhui Liu, and Jiaya Jia.
\newblock Spherical transformer for lidar-based 3d recognition.
\newblock In {\em Proceedings of the IEEE/CVF Conference on Computer Vision and Pattern Recognition}, pages 17545--17555, 2023.

\bibitem{Nuscenes}
Holger Caesar, Varun Bankiti, Alex~H. Lang, Sourabh Vora, Venice~Erin Liong, Qiang Xu, Anush Krishnan, Yu~Pan, Giancarlo Baldan, and Oscar Beijbom.
\newblock nuscenes: {A} multimodal dataset for autonomous driving.
\newblock {\em CoRR}, abs/1903.11027, 2019.

\bibitem{behley2019iccv}
J.~Behley, M.~Garbade, A.~Milioto, J.~Quenzel, S.~Behnke, C.~Stachniss, and J.~Gall.
\newblock {SemanticKITTI: A Dataset for Semantic Scene Understanding of LiDAR Sequences}.
\newblock In {\em Proc. of the IEEE/CVF International Conf.~on Computer Vision (ICCV)}, 2019.

\bibitem{ros}
{\em {Robot Operating System - Melodic Morenia}}, 2018.
\newblock {Available: \url{https://www.ros.org}}.

\bibitem{michaud2015towards}
Sebastien Michaud, Jean-Fran{\c{c}}ois Lalonde, and Philippe Giguere.
\newblock Towards characterizing the behavior of lidars in snowy conditions.
\newblock In {\em International Conference on Intelligent Robots and Systems (IROS)}, 2015.

\end{thebibliography}

\end{document}